%% file: main.tex
\pdfoutput=1

\documentclass[11pt]{article}

\usepackage[preprint]{acl}

\usepackage{times}
\usepackage{latexsym}

\usepackage[T1]{fontenc}

\usepackage[utf8]{inputenc}

\usepackage{microtype}

\usepackage{inconsolata}

\usepackage{graphicx}

\usepackage{amssymb}
\usepackage{pifont}
\usepackage{amsmath}
\usepackage{amsfonts}
\usepackage{subfigure}
\usepackage{graphicx}
\usepackage{booktabs}
\usepackage{url}
\usepackage{tabularx}
\usepackage{bigstrut}
\usepackage{multirow}
\usepackage{color}
\usepackage{xspace}

\usepackage{arydshln}
\usepackage{fontawesome}

\usepackage{tcolorbox}
\tcbuselibrary{raster}
\tcbuselibrary{breakable}
\usepackage{listings}
\input{macro}

\newcommand{\cmark}{\textcolor{green}{\ding{51}}}%
\newcommand{\xmark}{\textcolor{red}{\ding{55}}}

\newcommand{\method}{\textsc{OfficeBench}\xspace}

\title{\method: Benchmarking Language Agents across Multiple Applications for Office Automation}

\author{
\textbf{Zilong Wang}$^{1}$\quad 
\textbf{Yuedong Cui}$^{2}$\quad
\textbf{Li Zhong}$^{1}$\quad
\textbf{Zimin Zhang}$^{2}$\\
\textbf{Da Yin}$^{2}$\quad
\textbf{Bill Yuchen Lin}$^{3}$\quad
\textbf{Jingbo Shang}$^{1}$\\
$^{1}$UC San Diego\quad
$^{2}$UCLA\quad
$^{3}$Allen Institute for AI\\
\texttt{\{zlwang,lizhong,jshang\}@ucsd.edu}\\
\texttt{\{cui54,ziminz19\}@g.ucla.edu}\\
\texttt{da.yin@cs.ucla.edu}\quad
\texttt{yuchenl@allenai.org}
}

\begin{document}
\maketitle
\begin{abstract}
\input{contents/0-abstract}
\end{abstract}

\input{contents/1-introduction}
\input{contents/2-related_work}

\input{contents/3-framework}

\input{contents/3-settings}

\input{contents/4-experiments}

\input{contents/5-conclusion}

\bibliography{custom}

\clearpage
\appendix
\input{contents/X-appendix}

\end{document}

%% file: macro.tex
\makeatletter
\lst@InstallKeywords k{attributes}{attributestyle}\slshape{attributestyle}{}ld
\makeatother

\definecolor{pythonblue}{rgb}{0.16,0.12,0.93}
\definecolor{cppgreen}{rgb}{0.16,0.42,0.16}
\definecolor{promptinsert}{HTML}{bfefff}
\definecolor{compcolor}{HTML}{90EE90}
\definecolor{codehlcolor}{HTML}{ffec8b}
\definecolor{codehlcolor2}{HTML}{ffbbff}
\definecolor{bgcolor}{rgb}{0.95,0.95,0.92}
\definecolor{spblue}{HTML}{00b5ea}

\lstdefinestyle{python}{
    language=Python,
    basicstyle=\fontsize{8}{10}\ttfamily,
    keywordstyle=\color{blue},
    commentstyle=\color{gray},
    stringstyle=\color{black},
    showstringspaces=false,
    breaklines=true,
    breakindent=0pt,
    breakatwhitespace=false,
    escapeinside={(*@}{@*)}
}

\lstdefinestyle{cpp}{
    language=C++,
    basicstyle=\fontsize{8}{10}\ttfamily,
    keywordstyle=\color{blue},
    commentstyle=\color{gray},
    stringstyle=\color{green},
    showstringspaces=false,
    breaklines=true,
    breakindent=0pt,
    breakatwhitespace=false,
    escapeinside={(*@}{@*)}
}

\lstdefinestyle{plain}{
    basicstyle=\fontsize{8}{10}\ttfamily,
    keywordstyle=\color{blue},
    commentstyle=\color{gray},
    stringstyle=\color{green},
    showstringspaces=false,
    breaklines=true,
    breakatwhitespace=false,
    breakindent=0pt,
    escapeinside={(*@}{@*)},
    literate={á}{{\'a}}1 {ã}{{\~a}}1 {é}{{\'e}}1,
}

\lstdefinestyle{demo}{
    basicstyle=\fontsize{6}{7}\ttfamily,
    keywordstyle=\color{blue},
    commentstyle=\color{gray},
    stringstyle=\color{green},
    showstringspaces=false,
    breaklines=true,
    breakatwhitespace=false,
    breakindent=0pt,
    escapeinside={(*@}{@*)},
    literate={á}{{\'a}}1 {ã}{{\~a}}1 {é}{{\'e}}1,
}

\lstdefinestyle{example}{
    basicstyle=\fontsize{8}{10}\ttfamily,
    keywordstyle=\color{spblue}\bfseries\underline,
    commentstyle=\color{gray},
    stringstyle=\color{green},
    showstringspaces=false,
    breaklines=true,
    breakatwhitespace=false,
    breakindent=0pt,
    escapeinside={(*@}{@*)},
    morekeywords={ Question, Answer, Prediction, Results, Explanation },
}

\lstdefinestyle{python2}{
    language=Python,
    basicstyle=\fontsize{8}{10}\ttfamily,
    keywordstyle=\color{blue},
    commentstyle=\color{gray},
    stringstyle=\color{green},
    showstringspaces=false,
    breakatwhitespace=false,
    breaklines=true,
    breakindent=0pt,
    escapeinside={(*@}{@*)}
}

\lstdefinestyle{cpp2}{
    language=C++,
    basicstyle=\fontsize{8}{10}\ttfamily,
    keywordstyle=\color{blue},
    commentstyle=\color{gray},
    stringstyle=\color{green},
    showstringspaces=false,
    breaklines=true,
    breakindent=0pt,
    breakatwhitespace=false,
    escapeinside={(*@}{@*)}
}

\lstdefinestyle{sql}{
    language=SQL,
    basicstyle=\fontsize{8}{10}\ttfamily,
    keywordstyle=\color{blue},
    commentstyle=\color{green},
    stringstyle=\color{black},
    showstringspaces=false,
    breakatwhitespace=false,
    breaklines=true,
    breakindent=0pt,
    escapeinside={(*@}{@*)}
}

\lstdefinestyle{prompt}{
    language=Python,
    basicstyle=\fontsize{8}{10}\ttfamily,
    keywordstyle=\color{blue},
    commentstyle=\color{gray},
    showstringspaces=false,
    breaklines=true,
    keepspaces=true, 
    breakindent=0pt,
    breakatwhitespace=false,
    showspaces=false,   
    escapeinside={(*@}{@*)}
}
\lstdefinestyle{text}{
    basicstyle=\fontsize{8}{10}\ttfamily,
    showstringspaces=false,
    breaklines=true,
    breakatwhitespace=false,
    breakindent=0pt,
    keepspaces=true,
    showspaces=false,   
    escapeinside={(*@}{@*)}
}

%% file: contents/0-abstract.tex
Office automation significantly enhances human productivity by automatically finishing routine tasks in the workflow. Beyond the basic information extraction studied in much of the prior document AI literature, the office automation research should be extended to more realistic office tasks which require to integrate various information sources in the office system and produce outputs through a series of decision-making processes. We introduce \method, one of the first office automation benchmarks for evaluating current LLM agents' capability to address the office tasks in realistic office workflows. \method requires LLM agents to perform feasible long-horizon planning, proficiently switch between applications in a timely manner, and accurately ground their actions within a large combined action space, based on the contextual demands of the workflow. Applying our customized evaluation methods on each task, we find that GPT-4 Omni achieves the highest pass rate of 47.00\%, demonstrating a decent performance in handling office tasks. However, this is still far below the human performance and accuracy standards required by real-world office workflows. We further observe that most issues are related to operation redundancy and hallucinations, as well as limitations in switching between multiple applications, which may provide valuable insights for developing effective agent frameworks for office automation.\footnote{Code \& Data: \url{https://github.com/zlwang-cs/OfficeBench}}

%% file: contents/1-introduction.tex
\section{Introduction}

Office automation plays a pivotal role in interacting with diverse environments to accomplish complex goals set by users. In the rapidly evolving landscape of workplace technology, the integration of office automation into daily tasks represents a critical advancement with the potential to significantly enhance human efficiency and transform traditional workflows~\cite{aghion2023effects,filippi2023automation}. By automating routine and time-consuming tasks, office automation systems free up human workers to focus on more strategic and creative aspects of their roles~\citep{howcroft2023automation}.

\begin{figure}[t]
    \centering
    \includegraphics[width=0.95\linewidth]{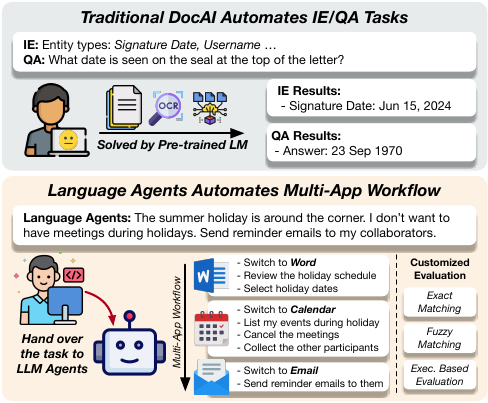}
    \caption{\textbf{\method is one of the first office automation benchmarks for language agents.} We assess the ability of language agents to perform complex office workflows across multiple applications using customized evaluation methods, such as Exact Matching, Fuzzy Matching, and Execution-based Evaluation.}
    \vspace{-4mm}
    \label{fig:teaser}
\end{figure}

Towards the ambitious goal of automating office work, numerous efforts have been made from both industry and academia~\citep{binmakhashen2019document,cui2021document}. One common direction is Document AI which automates the fundamental tasks, such as information extraction and question answering, by pre-trained language models~\citep{xu2020layoutlm, wang2020docstruct, wang2021layoutreader, garncarek2021lambert, xu2021layoutlmv2, wang2022mgdoc, wang2022towards, perot2023lmdx, nguyen2024docmaster}. Following this direction, many benchmarks include structured documents with detailed annotations, requiring language models to understand the rich structure and extract the key information or respond to the specific questions posed within these documents.~\citep{jaume2019funsd, park2019cord, mathew2021docvqa, wang2023vrdu}. 

\begin{figure*}
    \centering
    \includegraphics[width=0.95\linewidth]{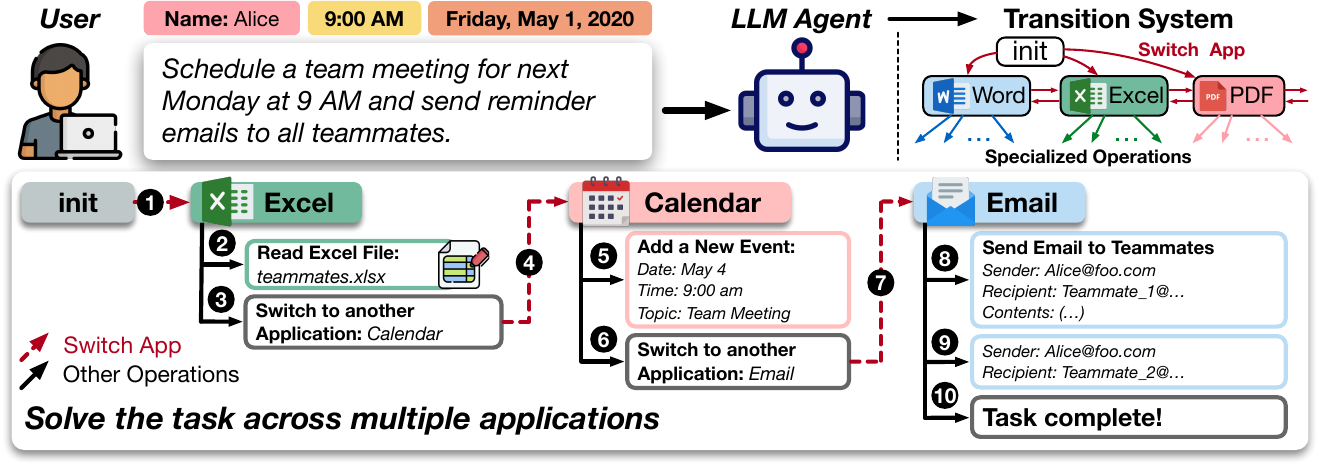}
    \vspace{-3mm}
    \caption{\textbf{Illustration of the workflow in \method{}} where the LLM agent leverages the operations from multiple applications to systematically construct an operation chain that addresses the office tasks effectively. The framework is formulated as a transition system where the current application serves as the \textit{state} and the operations serve as the \textit{transitions}. Specialized operations, such as \texttt{read\_file} and \texttt{send\_email}, perform specific tasks.} 
    \label{fig:officebench}
\end{figure*}

However, a realistic office environment extends far beyond basic extraction tasks. Prior works on structured document understanding are only part of the office automation pipeline. For example, extracting data from PDF invoices is just the beginning; the full process involves integrating this data into financial software, flagging discrepancies, and generating payment reminders. It is necessary to develop and evaluate an entire office automation framework that seamlessly integrates various information within the office system, ensuring the output aligns with logical planning processes. With the impressive planning and reasoning capabilities of large language models (LLMs)~\citep{achiam2023gpt,team2023gemini,reid2024gemini}, language agents powered by them are expected to construct feasible operation chains to execute the typical workflows for human labors, including but not limited to the information extraction tasks. 

To this end, we propose \textbf{\method}, as one of the first office automation evaluation benchmarks for LLM agents. By deploying agents within simulated human labor workflows that replicate the complexity and variability of modern office environments, this benchmark is instrumental in assessing the ability of language agents to handle a variety of office tasks across different applications.
The \method benchmark operates within a Docker environment pre-installed with office applications such as \texttt{Word}, \texttt{Excel}, \texttt{calendar}, and \texttt{email clients} to simulate various scenarios, including sending emails, editing tables, and scheduling events.
With the consideration of multiple applications, LLM agents are required to demonstrate their proficiency in \textit{switching} between applications on time, and grounding their actions accurately from a \textit{large combined action space} based on the contextual demands of the workflow. 
Furthermore, \method incorporates various evaluation methods including exact matching, fuzzy matching, and execution-based evaluation customized for each test example in the benchmark, allowing for the results of agent actions to be assessed in various file formats. This feature is critical in precisely assessing the agent ability to follow the user task instruction individually.

We extensively evaluate state-of-the-art LLMs as language agents in following natural language commands and perform various office tasks across multiple applications. We evaluate the proprietary GPT-3.5, GPT-4~\citep{achiam2023gpt}, Gemini-1.0~\cite{team2023gemini}, Gemini-1.5~\citep{reid2024gemini}, and open-weights Llama 3~\citep{llama3}, Qwen 2~\citep{qwen}. 
The experimental results indicate that GPT-4 Omni achieves the highest pass rate of 47.00\%, showcasing a decent performance of current LLMs in handling office automation tasks. However, this is still well below the accuracy standards required by real-world office workflows, highlighting the need for continued research to further explore the limits of language agents powered by LLMs.
We further conduct ablation study on application switching in multi-application scenarios, and analyze the failure cases. We identify issues related to operation redundancy and hallucinations, as well as limitations of current LLM in complex planning across multiple applications. 

With our proposed \method benchmark, we would like to shed new light on more robust and effective language agents, facilitating the development of advanced automation for realistic, everyday tasks, and breaking down invisible barriers in modern workspace, including those related to disability, education, and cultural differences. 

%% file: contents/2-related_work.tex
\section{Related Work}

\paragraph{Language Agent Benchmarks}
Previous studies typically assess LLM agents in focused domains, such as arithmetic reasoning, which targets correct solutions, and tool-use, evaluating agents' proficiency in employing tools~\cite{yang2018hotpotqa,cobbe2021training,xu2023rewoo,liu2023agentbench,wang2023mint,ma2024agentboard,zhong2024can,zhong2024ldb}. The most recent evaluation benchmarks have increasingly focused on real-world scenarios, including web and OS environments~\cite{deng2024mind2web,zhou2023webarena,koh2024visualwebarena,lu2024weblinx,xie2024osworld}, wheere they enable agents to interact with web/OS interfaces using keyboard and mouse actions. Different from these prior works, \method is an agent evaluation benchmark specifically designed to assess LLM abilities within real-world workflows, requiring the operation of multiple software applications to complete tasks. \method encompasses a larger action space and demands LLMs to possess the capability in switching between software applications as needed. It is also one of the first benchmarks to offer customized evaluation methods tailored to different software and individual tasks, ensuring a precise assessment. \method provides an extensible and cost-effective evaluation platform, supporting the addition of new applications and tasks compatible with the Bash environment with less manual effort than the complex simulators in OSWorld annotated with around 1800 human hours~\cite{xie2024osworld}.

\paragraph{Document AI Benchmarks} 
Document AI focuses on various structured documents, including invoices, receipts, forms, and tables. Previous studies primarily focus on the information extraction tasks on these documents, assessing the capability of language models in understanding the textual contents and rich structural information. CORD~\citep{park2019cord} and SPOIE~\citep{huang2019icdar2019}, FUNSD~\citep{jaume2019funsd} and VRDU~\citep{wang2023vrdu} incorporate grocery receipts or multi-domain forms for entity extraction tasks. DocVQA~\citep{mathew2021docvqa} formulates the structured document understanding as an extractive QA task. Realistic office scenarios involve more comprehensive workflows with multiple applications. The information extraction or question answering tasks are only parts of the complex workflow. Our proposed \method go beyond the document-based benchmarks and evaluate the powerful LLM agents in calling different applications for general office automation.

We further compare \method with recent benchmarks in Document AI and language agents from different perspectives in Appendix~\ref{app:benchmark-comparison}. \method excels in cross-application scenarios, offering a diverse suite of precisely curated customized evaluation functions for each task. Additionally, it supports a larger action space and provides more extensible task annotation and environment creation capabilities.

%% file: contents/3-framework.tex
\section{\method: Modeling Office Works Across Multiple Applications}\label{sec:env}

Office automation must be capable of complex planning and reasoning to construct an applicable chain of actions for solving real-world tasks. While LLMs have demonstrated satisfactory performance in single-application scenarios~\citep{wang2023chain,zhou2023webarena,deng2024mind2web,lu2024weblinx}, comprehending the diverse execution environments including various applications and effectively managing a vast action space for realistic tasks remains a challenge. To evaluate LLM agent performance on office automation in multi-application scenarios, in \method benchmark, we develop a realistic and extensible framework designed to simulate office work scenarios which incorporates applications such as \texttt{Word}, \texttt{Excel}, \texttt{PDF}, \texttt{Shell} and \texttt{email client}. The framework also supports a large set of valid actions applicable to these applications. LLM agents should smartly leverage the applications supported in the environment with the valid actions by utilizing their advanced planning and reasoning abilities. In this section, we present the overall framework of \method, detailing the multi-application environment and the workflow of the automation system with the large action space.

\input{tables/app_action_info}

\subsection{Multi-Application Environment} 

We formulate an office task for autonomous agents as a task description $T$ with a variety of applications commonly used in office scenarios, such as \texttt{Word}, \texttt{Excel}, \texttt{PDF}, and \texttt{Shell}. Each application is defined by a distinct set of operations tailored to the specific usage. These operations are denoted as $A = \{\alpha_1,...,\alpha_n\}$, where $A$ represents the application, and each $\alpha_i$ represents an individual operation within this application's environment. For example, in the context of the \texttt{Excel} application, we design specialized operations such as \texttt{read\_excel\_file} and \texttt{set\_cell\_content}, which are explicitly designed to interact with spreadsheet data.

Overall, as shown in Table~\ref{tab:app-action-info}, we have designed a total of 9 applications within the multi-application environment of \method to simulate a realistic office work scenario. These include specialized applications, such as \texttt{Word},
\texttt{Excel},
\texttt{PDF},
\texttt{Calendar},
\texttt{Email},
\texttt{OCR},
\texttt{ChatGPT},
\texttt{Shell}, and an auxiliary application \texttt{System}. We develop various basic operations for each application as listed in Table~\ref{tab:app-action-info}. In \method, the LLM agents are able to leverage the operations from multiple applications to systematically construct an operation chain that addresses the office tasks effectively.

\paragraph{Application Transition}
In a single-application environment, it is relatively straightforward to consistently engage with only one application, querying the LLM agents for subsequent actions based on interaction feedback with that application. However, when it comes to a multi-application environment, it is necessary to design approaches to coordinate the various applications. Drawing inspiration from the idea of operating systems, we introduce an auxiliary application named \texttt{System}, serving as a foundational platform that coordinates other specialized applications. This \texttt{System} application is crucial as it includes the operation \texttt{switch\_app}, which is designed to manage the seamless transition between multiple execution environments tailored for various applications. 
Once the agent self-identifies that it has already obtained what it expects from an application, then it can use the \texttt{switch\_app} operation to change to another one.
For example, when solving a task ``\textit{Send emails to the participants of the meeting today.}'', an LLM agent needs to switch to \texttt{Email} after acquiring participants information from the \texttt{Calendar}.

\paragraph{Operation Observation}
We integrate the operation outputs into the observation space of the LLM agents, formatting these outputs textually so that the agents can directly learn from the rich signals they contain. Given the variety of operations across different applications in \method, we handle each case individually.
In simpler cases, we directly print the outputs for LLM agents. For example, when calling \texttt{run\_command} with the \texttt{Shell} application, we simply copy the terminal outputs to the execution history. In cases involving structured data,  we decode the structure and retain the essential information in the textual outputs. For example, when calling \texttt{read\_excel\_file} with the \texttt{Excel} application, we list the cell values along with their indices in the format $(i, j): \texttt{Value}$ where $i, j$ are the row and column indices, and \texttt{Value} is the content of the specified cell. Refer to Appendix~\ref{app:observation} for the formalized outputs in more details.

\subsection{Autonomous Workflow~\label{sec:workflow}}
Based on the multi-application environment and supported action space of \method, we formulate the autonomous workflow as a transition system $\mathcal{E}=\{\mathcal{S}, \mathcal{A}, \mathcal{O}, \mathcal{T}\}$, standing for state space $\mathcal{S}$, action space $\mathcal{A}$, observation space $\mathcal{O}$, and transition function $\mathcal{T}: \mathcal{S}\times \mathcal{A} \to \mathcal{S}$, as shown in Figure~\ref{fig:officebench}. We set the currently selected application as the \textit{state} in the transition system and introduce the restricted action space for each application. We further specify the observation space and termination condition of the agent system in this section. 

\paragraph{Restricted Action Space} 
The current application $A$ in use determines the set of actions that are currently valid. Given the extensive range of operations across various applications, we restrict the action space to the specialized operations within $A$. Additionally, we include the \texttt{switch\_app} and \texttt{submit} operation in the action space, allowing the LLM agent to switch to another application or submit the task when necessary. Specifically, supposing the application at timestamp $t$ is $A_t$, the action space is formulated as,
$
    \{\alpha_{t1}, \alpha_{t2}, ..., \alpha_{tm}\} \cup \{\texttt{switch\_app}, \texttt{submit}\}
$,
where $\{\alpha_{t1}, \alpha_{t2}, ..., \alpha_{tm}\}$ are valid actions under the application $A_t$.

\paragraph{Observation Space}
In \method, we provide the LLM agents with the full execution history in the prompt as the observation so the LLM agents can determine the next action based on the previous actions and their observed outputs, leading the system to transition to the next state. Specifically, at timestamp $t$, the execution history of previous actions is represented as
$
    H_t = [(A_1, \alpha_1, o_1),..., (A_{t-1}, \alpha_{t-1}, o_{t-1})]
$,
where $(A_{i}, \alpha_{i}, o_{i})$ denotes the application, action, and the observed outputs at timestamp $i$, respectively. The observed outputs of each action are introduced in Section~\ref{sec:env} and listed in Table~\ref{tab:app-action-info-detail}.
The LLM agent predicts the next action among the restricted action space based on the history $H_t$ and triggers a transition function to proceed to the next state.

\paragraph{Termination Condition}
LLM agents are designed to iteratively predict and execute operations as a transition system until the given task is completed. In the \texttt{System} application, we implement an operation, \texttt{submit\_task}, as a symbol of normal termination. Nevertheless, due to the limitations of current LLM agents, we have identified two additional conditions that necessitate terminating the agent system prematurely -- \textit{Operation Stagnation:} If an LLM agent continuously generates the same operation multiple times, we interpret this as a failure. Specifically, if this repetition occurs 5 times consecutively, we terminate the system and classify the task as unsuccessful in \method. 
\textit{Iteration Overflow:} Given the constraints on resources, it is necessary to limit the number of iterations an LLM agent can perform. Therefore, we set a maximum step limit as 50 to prevent excessive resource use and ensure timely task completion.

\subsection{Implementation Details}\label{sec:folder}

We build \method in a Docker environment with pre-installed applications and use Python libraries to automate the operations. We create a file system for the documents, emails, and calendar events involved in the tasks. We formulate each user's emails as ".eml" files under a specific directory (e.g. \texttt{/emails/[username]/}). 
Similarly, we save the user's calendar events as a ".ics" file (e.g. \texttt{/calendar/[username].ics}). We save the other ordinary documents in \texttt{/data/}. After the agents finish the task, we save the entire file system and run customized evaluation to check the correctness. Refer to Appendix~\ref{app:prompt} for prompt examples used in our experiments.

%% file: tables/app_action_info.tex
\begin{table}[t]
  \centering
  \small
  \resizebox{\linewidth}{!}{
    \setlength{\tabcolsep}{2mm}{
    \begin{tabular}{ll}
    \toprule
    \textbf{Applications} & \textbf{Operation Examples} \\
    \midrule
    \texttt{System} (2) & \texttt{switch\_app, submit\_task} \\
    \texttt{Word} (4) & \texttt{convert\_to\_pdf, write\_to\_file} \\
    \texttt{Excel} (5) & \texttt{set\_cell, read\_excel\_file} \\
    \texttt{PDF} (3) & \texttt{convert\_to\_doc, read\_pdf\_file} \\
    \texttt{Calendar} (3) & \texttt{create\_event, delete\_event} \\
    \texttt{Email} (3) & \texttt{list\_emails, send\_email} \\
    \texttt{OCR} (1) & \texttt{recognize\_text} \\
    \texttt{ChatGPT} (1) & \texttt{query\_chatgpt} \\
    \texttt{Shell} (1) & \texttt{run\_command} \\
    \bottomrule
    \end{tabular}%
  }
  }
  \caption{\textbf{Applications and their corresponding operations} implemented in \method for simulating a realistic office scenario. The number in the brackets are the total number of the operations in this application. See Appendix~\ref{app:all-operations} for details.}
  \label{tab:app-action-info}%
\end{table}%

%% file: contents/3-settings.tex
\section{Benchmark Annotation and Evaluation}

In \method, we construct a comprehensive suite of 300 tasks to evaluate the performance of LLM agents in office automation. For each task, we synthesize documents, emails, and calendar events involved in the tasks to simulate a realistic scenario. 
We also design customized evaluation methods, including the exact and fuzzy matching, and the execution-based evaluation. We outline the annotation process and describe our comprehensive evaluation framework in this section.

\subsection{Task Annotation}
\method evaluates the capability of LLM agents in managing multiple applications with the three categories of tasks, Single App, Two Apps, and Three Apps, specifying the number of involved apps. Among these tasks, the difficulty level increases with more applications involved. Overall, we annotate 93, 95, and 112 tasks in these three categories, respectively.

\paragraph{Single App Tasks}
In the Single App category, tasks are relatively easier. The LLM agents select one application in the beginning, adhere to it, and plan an operation chain to solve the task. With these simpler tasks, we aim to investigate whether the LLM agents are able to understand the functionality of the elementary operations in each application. 

\paragraph{Two Apps Tasks} In the other two categories, Two/Three Apps, LLM agents need to switch to another application once they self-identify that they have already obtained what it expects from the current application. When annotating tasks in the Two Apps category, we request annotators to brainstorm realistic and diverse tasks relevant to every pair of applications in \method. For example, when integrating \texttt{PDF} and \texttt{Email}, we design a task ``\textit{Extract a notification from a business travel image and send emails to Bob and Tom}''.

\paragraph{Three Apps Tasks} In order to further evaluate LLM agents with more challenging tasks, we expand the tasks in the Two Apps category with one more relevant application while ensuring the validity of the combination. In this way, we annotate more complex tasks in the Three Apps category. For example, we already annotate a task of Two Apps (\texttt{Excel} and \texttt{Calendar}): ``\textit{Schedule a team training session for all participants from the Excel file and create calendar events for each member}''. We add a relevant application \texttt{Email}, requesting LLM agents to email the training details to each participant in the following steps. Despite the seemingly simple addition, the tasks in the Three Apps category present a greater challenge to LLM agents, requiring them to adeptly manage dynamic switching between the applications.

\subsection{Data Synthesis}

We aim to simulate a realistic office environment in \method. A delicate file system is indispensable. We synthesize the documents of various formats, emails, and calendar events for each of the tasks in our benchmark. To eliminate the sensitive privacy issues, we resort to ChatGPT\footnote{\url{https://chatgpt.com/}} and random generators instead of using real data. Specifically, we query ChatGPT to generate paragraphs on specific topics as needed, and run Python programs to generate random numbers. For example, to synthesize exam scores for a class, we initially query ChatGPT to generate a list of common student names and then assign each student a random score ranging from 0 to 100. When it comes to files with special formats, such as images, PDFs, we use the HTML format as an intermediary. In particular, we first edit the HTML files to involve rich layout structure and then convert it to the requested formats. Similarly, for emails and calendar events, we fill in the fields in the special data structure with synthesize contents. Finally, we copy the involved data to the \method Docker environment for the evaluation of the LLM agents.

\input{tables/results}

\subsection{Evaluation Framework\label{sec:evaluation-framework}}

To evaluate LLM agents within the simulated office workflow of \method, it is crucial to develop a precise and reliable method for assessing the output files produced by these agents after planning and execution. Given the diversity of the office work tasks, the task metrics may greatly vary due to the different task requirements and involved applications. Following \citet{zhou2023webarena, xie2024osworld}, we incorporate the exact matching, fuzzy matching, and execution-based methods into the evaluator of \method (See Appendix~\ref{app:evaluation} for detailed examples).

\paragraph{Exact Matching \& Fuzzy Matching}
In the exact matching setting, we utilize our annotated ground-truth outputs of the tasks as references and assess whether LLM agent's final outputs match them exactly. For example, given a task ``\textit{Bob got 98 points in the final exam. Add his score in final-exam.xlsx.}'', we add a new row for Bob and his score in the specified file, \textit{final-exam.xlsx}, and compare the file processed by the LLM agent with the ground-truth annotation. However, when evaluating more complex tasks, it becomes challenging to design strict criteria for the correct answer. For example, consider the task: ``\textit{Add a meeting to Bob's calendar from 10:30 am to 11:00 am tomorrow.}'' In this case, we employ a fuzzy matching function to assess accuracy. This function checks the correctness of the timestamps in the calendar event and verifies that the event subject includes the keyword \textit{meeting}. We disregard other details of the event, adopting a more flexible criterion for correctness.

\paragraph{Execution-based Evaluation} In addition to exact and fuzzy matching, we incorporate execution-based evaluation methods to address more complicated scenarios. Specifically, we run a short code snippet to verify the correctness of results from the LLM agent since the expected results are not unique. Take the task ``\textit{Set up a meeting for Alice and Bob tomorrow when they are both free.}'' as an example. This requires the LLM agent to check Alice's and Bob's schedules to pinpoint a mutually available time slot. To validate the result, we implement a code snippet that checks if the meeting is scheduled in both Alice's and Bob's calendars and ensures there are no overlapping commitments or time conflicts.

%% file: tables/results.tex
\begin{table*}[t]
  \centering
  \resizebox{\linewidth}{!}{
    \setlength{\tabcolsep}{2.5mm}{
    \small
    \begin{tabular}{lcccc}
    \toprule
{\textbf{LLM Agents}} & {\textbf{Single App (93)}} & {\textbf{Two Apps (95)}} & {\textbf{Three Apps (112)}} & {\textbf{Overall (300)}} \\
\midrule
    \textbf{\textit{Proprietary Models}} &       &       &       &  \\

    \quad Gemni-1.0 Pro \scriptsize{(Latest update: Feb 2024)} & {24.73} & {13.68} & {0.89} & {12.33} \\

    \quad Gemni-1.5 Flash \scriptsize{(Latest update: May 2024)} & {34.41} & {24.21} & {0.89} & {18.67} \\

    \quad Gemni-1.5 Pro \scriptsize{(Latest update: May 2024)} & {41.94} & {32.63} & {7.14} & {26.00} \\

    \hdashline

    \quad GPT-3.5 Turbo \scriptsize{(0125)} & {8.60} & {7.45} & {0.89} & {5.35} \\
    
    \quad GPT-4 Turbo \scriptsize{(2024-04-09)}  & {56.99} & {50.63} & {11.61} & {38.00} \\

    \quad GPT-4 Omni \scriptsize{(2024-05-13)} & {\textbf{64.52}} & {\textbf{60.00}} & {\textbf{21.43}} & {\textbf{47.00}} \\
    
    \midrule
    \textbf{\textit{Open-weight Models}} &       &       &       &  \\
    \quad Llama 3 \scriptsize{(70B-Instruct)} &  \textbf{39.79}  &  \textbf{41.05}  & 5.36  &  \textbf{27.33}  \\
    \quad Qwen 2 \scriptsize{(72B-Instruct)} &    30.23   &  28.42  &  \textbf{8.04}  &  21.16 \\

    \midrule
    Human Performance & 96.00 & 96.00 & 88.00 & 93.33 \\
    
    \bottomrule
    \end{tabular}%
    }
    }
  \caption{\textbf{Pass rates (\%) on agent automation tasks} from \method for the proprietary models, Gemini-1.0, Gemini-1.5, GPT-3.5, GPT-4, and the open-sourced models, Llama 3 and Qwen 2. We divide the tasks into ``Single/Two/Three App(s)'', specifying the number of applications required by the tasks; we also report the overall performance; the number in the brackets denotes the number of tasks in each subcategory.
  \textbf{Bold} denotes the best performance among the proprietary or the open-weight models. 
  }
  \vspace{-3mm}
  \label{tab:results}%
\end{table*}%

%% file: contents/4-experiments.tex
\section{Experiments}

With our proposed \method, we evaluate the office automation capability of the proprietary LLMs, including Gemini-1.0~\citep{team2023gemini}, Gemini-1.5~\citep{reid2024gemini}, GPT-3.5~\citep{achiam2023gpt}, and GPT-4~\citep{achiam2023gpt}, and the open-weights LLMs, including Llama 3~\citep{llama3} and Qwen 2~\citep{qwen}, as these models are among the highest-ranking LLMs available~\cite{open-llm-leaderboard}. We also ask two computer science graduate students to perform these task and report the human performance (See Appendix~\ref{app:human} for error analysis for human annotators).

In \method, the LLM agents need to interact with the multiple applications available in the environment, construct a feasible operation chain, and accomplish the task step by step. We adopt the end-to-end prompting approach to guide LLMs in planning and executing workflows autonomously, without the need for manually selected demonstrations. 
In this way, we eliminate the biases introduced by the cherry-picking demonstrations and guarantee the reliability and robustness of the experimental results on \method. 
We leverage our designed customized evaluation methods discussed in Section~\ref{sec:evaluation-framework} for each test task to verify if the outcomes from the LLM agents pass. We use \textit{pass rate}, $\frac{\# \text{Pass Examples}}{ \# \text{All Examples}}$, as our final metrics.

\subsection{Main Results}

We demonstrate the experimental results of the LLM agents in Table~\ref{tab:results}. We present both the overall performance and fine-grained performance of the evaluated LLM agents across the subcategories of ``Single/Two/Three App(s)''. We separate the LLMs into two groups: proprietary models and open-weights models. Within each group, the best-performing model is highlighted in bold. 
Table~\ref{tab:results} shows that GPT-4 Omni and Llama 3 lead their respective groups, achieving overall pass rates of 47.00\% and 27.33\% for proprietary and open-weights models, respectively. These decent results show the basic capability of current LLM agents in solving office automation tasks, while there is still a huge gap to the human performance.  We also observe that the open-weight Llama 3 even surpasses the proprietary Gemini-1.5, underlining that open-weight models are not necessarily worse than the proprietary models. 

Specifically, we observe that performance diminishes greatly when tasks require interactions between multiple applications, underscoring the inherent complexities associated with more intricate tasks. The state-of-the-art LLM agent, GPT-4 Omni, can only achieve 21.43\% in the subcategory of ``Three Apps'', indicating a dramatic performance drop compared with ``Two Apps'' and ``Single App'' subcategories. We attribute this tendency to the limited capability of LLMs in tackling complex workflows with multiple applications, including the data formats specific to each application and the planning with different applications. Refer to Section~\ref{sec:error} for the detailed error analysis.

\input{tables/ablation}

\begin{figure*}[t]
    \centering
    \includegraphics[width=0.98\linewidth]{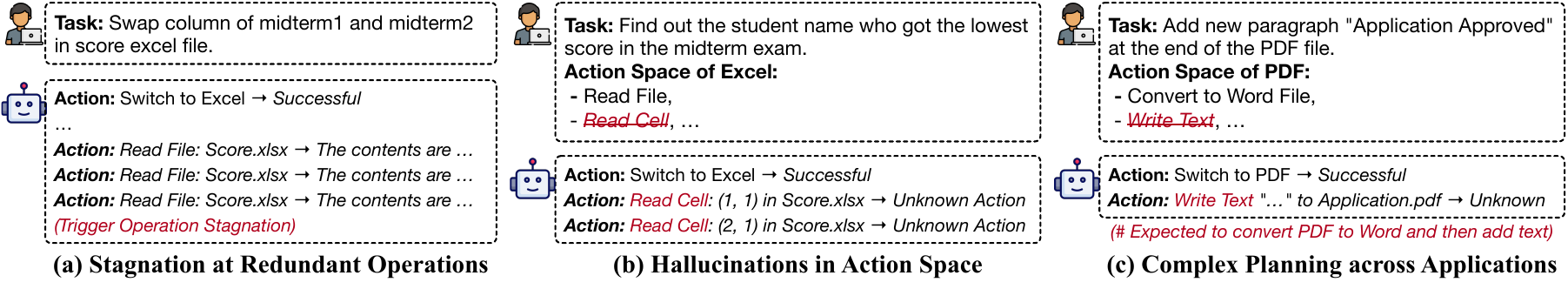}
    \caption{\textbf{Typical failure cases of the LLM agents} when solving office automation tasks in \method. We highlight the repeated redundant operations in (a), the hallucinated actions in (b), and the planning failure in (c). We omit the other contexts in the prompts and responses due to space limitation. The contents on the right side of the arrow "$\to$" are the observation of the action.}
    \vspace{-3mm}
    \label{fig:error-example}
\end{figure*}

\subsection{Ablation Study for Application Switching}

We highlight the complex workflows with multiple applications in our proposed \method which can well simulate the realistic office environment and investigate the planning and reasoning capabilities of LLMs in the complex workflow. 
In our designed framework, the LLM agents can switch between different applications by calling the \texttt{switch\_app} operation and get access to the action space specific to the target application. We denote this method as \textit{Use App Switch}.

In the ablation study, we follow the vanilla prompting method which lists all the operations regardless of the corresponding applications in the prompt. We denote this method as \textit{List All Operations}. We investigate the performance of GPT-4 Omni and Llama 3 under these two settings as they are the top-performing proprietary and open-weights models in \method, respectively. 

We report the performance and also calculate the average number of tokens used per iteration in cases that terminates normally\footnote{We exclude the cases that terminate due to \textit{Operation Stagnation} or \textit{Iteration Overflow}, which introduces meaningless wasted tokens.} in Table~\ref{tab:ablation}.
We observe that the application switching mechanism outperforms its counterpart, enabling LLM agents to effectively manage multiple applications within complex workflows. 
This enhancement can be attributed to the more concise natural language instructions and the constrained action space in the prompts. The action space of next step is largely constrained to the operations of the current application via the application switching operation (refer to Section~\ref{sec:workflow} for details).

\subsection{Error Analysis}\label{sec:error}

We further conduct error analysis on the outcomes from the LLM agents and summarize the typical failure cases in Figure~\ref{sec:error}.

\paragraph{Stagnation at Redundant Operations} As illustrated in Figure~\ref{sec:error} (a), although the activation of the \texttt{read\_file} operation to examine the spreadsheet's contents is initially successful, the LLM agent persistently repeats this operation. This occurs despite the feedback provided from previous actions, leading to an operational stagnation.

\paragraph{Hallucinations in Action Prediction} LLM agents are susceptible to hallucinating actions not pre-defined in the given action space. As illustrated in Figure~\ref{sec:error} (b), we dynamically limit the action space to include only the operations pertinent to the currently selected application (see Section~\ref{sec:workflow}). However, under such a narrowed subset of the entire action space, we still frequently observe that LLM agents tend to hallucinate non-existent actions, resulting in non-executable commands. These malformed actions not only fail to achieve the expected outcomes but also lead to a significant API calling or local inference costs.

\paragraph{Complex Planning across Applications} In addition to the hallucinations discussed earlier, another type of non-executable actions can occur when LLM agents are tasked with complex workflows involving multiple applications. As shown in Figure~\ref{sec:error} (c), LLM agents are instructed to edit a PDF file. However, due to a lack of knowledge that editing a \texttt{PDF} file typically involves first converting it to a \texttt{Word} document, making the necessary edits, and then converting it back to \texttt{PDF}, the agents mistakenly attempt direct edits on the \texttt{PDF}. This step is beyond the pre-defined action space, thereby resulting in a malformed action error.

%% file: tables/ablation.tex
\begin{table}[t]
  \centering
    \resizebox{\linewidth}{!}{
    \setlength{\tabcolsep}{0.8mm}{
    \small
\begin{tabular}{lcccc}
\toprule
\textbf{LLM Agents} & \textbf{Single} & \textbf{Multiple} & \textbf{Overall $\uparrow$} & \textbf{\# Token $\downarrow$} \\
\midrule
\textbf{GPT-4O \scriptsize{(2024-05-13)}} &       &       &       &  \\
\  - \textit{Use App Switch} & \textbf{64.52} & \textbf{39.13} & \textbf{47.00} & \textbf{1439.82} \\
\  - \textit{List All Operations} & 63.44 & 35.75 & 44.33 & 2177.51 \\
\midrule
\textbf{Llama 3 \scriptsize{(70B-Instruct)}} &       &       &       &  \\
\  - \textit{Use App Switch} & \textbf{39.79} & 21.74  & \textbf{27.33} & \textbf{1181.28} \\
\  - \textit{List All Operations} & 29.03 & \textbf{24.15}  & 25.57  & 1630.15 \\
\bottomrule
\end{tabular}%
    }
    }
\caption{\textbf{Evaluation results (\%) of the ablation study} for application switching on \method. We investigate the performance of GPT-4 Omni and Llama 3 when using the \texttt{switch\_app} operation (\textit{Use App Switch}) or listing all operations in the prompt (\textit{List All Operations}) to manage the environment with multiple applications.}
\vspace{-4mm}
  \label{tab:ablation}%
\end{table}%

%% file: contents/5-conclusion.tex
\section{Conclusion}

We propose \method, one of the first office automation benchmarks for language agents. We simulate a realistic execution environment and extensively evaluate the capability of current powerful LLM agents in solving tasks across different applications. 
Our findings highlight the efficacy of application switching in managing operations from multiple applications, and identify the limitations of LLMs in tackling cross-application workflows. With \method, we aim to to advance the development of more robust and effective language agents for comprehensive office automation.

\section*{Limitation}
In this paper, we propose \method as one of the first office automation benchmarks for language agents. While the system comprehensively analyzes the capability of current LLMs in planning complex workflow involving multiple applications in office automation, we anticipate that a wider range of applications will further expand our benchmark's usage in more application scenarios. Additionally, we are exploring the potential of instruction tuning for language models specifically tailored to office automation tasks, aiming to boost their performance on \method.

\section*{Ethical Statement}

In our proposed \method benchmark, we only incorporate synthesized data in the file systems and all names of individuals and companies are fictitious and generated by ChatGPT. Therefore, we do not anticipate any major ethical concerns.

%% file: contents/X-appendix.tex
\section*{Appendix}

\input{tables/benchmark_comparison}

\section{Comparison with Recent Benchmarks~\label{app:benchmark-comparison}}
As shown in Table~\ref{tab:benchmark-comparisonn}, \method excels in cross-application scenarios, offering a diverse suite of precisely curated customized evaluation functions for each task. Additionally, it supports a larger action space and provides more extensible task annotation and environment creation capabilities.

\section{Applications and Operations~\label{app:all-operations}}

We list all the applications and their corresponding operations in Table~\ref{tab:app-action-info-detail}. We simulate a realistic execution environment for evaluating LLM agents in office automation tasks.

\section{Observation Formats~\label{app:observation}}
\begin{figure*}
    \centering
    \includegraphics[width=0.8\linewidth]{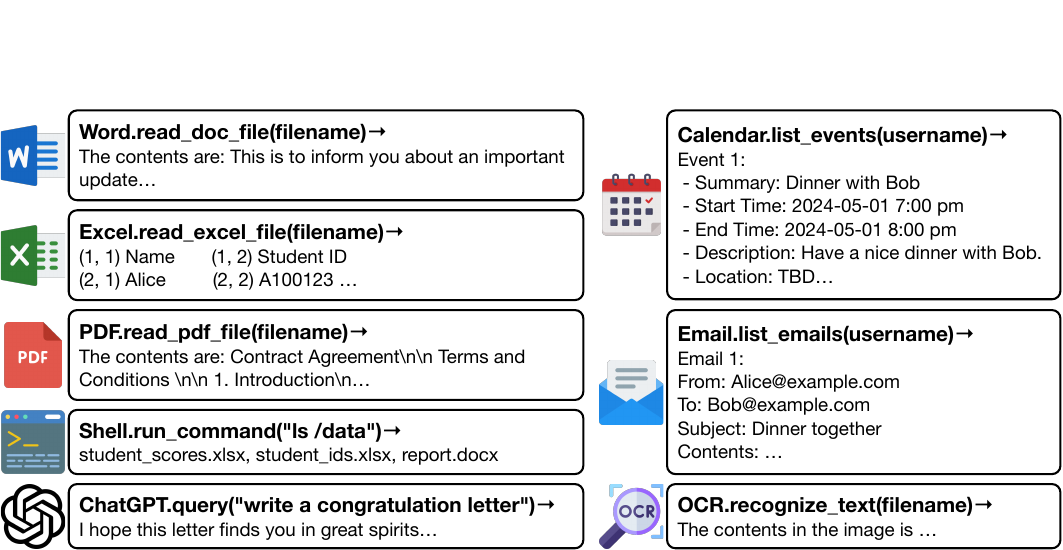}
    \caption{\textbf{Observation formats of representative operations} implemented in \method.}
    \label{fig:output-example}
\end{figure*}

We illustrate the observation formats of the representative operations in \method in Figure~\ref{fig:output-example}.

\section{Evaluation Methods~\label{app:evaluation}}
We provide more examples of evaluation methods used in \method in Table~\ref{tab:evaluation-method}.

\section{\method Prompts~\label{app:prompt}}
We provide prompt examples used in our experiments.
\begin{itemize}
    \item Prompt for application switching in Figure~\ref{fig:prompt-app-switch}.
    \item Prompt for planning next operation based on the trajectory in Figure~\ref{fig:prompt-plan}.
    \item Prompt of \textit{List All Operations} used in the ablation study in Figure~\ref{fig:prompt-ablation}.
\end{itemize}

\section{Error Analysis for Human Annotators~\label{app:human}}

The errors by human annotators mostly come from the misunderstanding of the task description or negligence in operations. For example, when solving ``\textit{Bob was invited to party hold by Jane Doe. Please send an email from Jane to Bob to notify Bob, and make a poster welcome.jpg for Bob}'', one annotator ignored the email sending requests and only created the poster. Another example is the task ``\textit{How many quarters did Bob win a scholarship? A scholarship is awarded only when a student's GPA exceeds 3.9.}'', where one annotator miscounted the number of quarters.

\input{tables/app_action_info_details}

\input{tables/evaluation_methods}

\input{prompts/first_app_switch}
\input{prompts/mid_prompt}
\input{prompts/ablation_prompt}

%% file: tables/benchmark_comparison.tex
\begin{table*}[t]
\centering
\resizebox{\linewidth}{!}{
\setlength{\tabcolsep}{1.8mm}{
\small
\begin{tabular}{lcccccc}
\toprule
      
\multirow{2}[1]{*}{\textbf{Benchmarks}} 
& \textbf{Office} & \textbf{\# Supported} & \multirow{2}[1]{*}{\textbf{Planning}}  & \textbf{Cross} & \multirow{2}[1]{*}{\textbf{Extensible}} & \textbf{Customized} \\ 
& \textbf{Automation} & \textbf{Actions} &  & \textbf{App.} &  &  \textbf{Task Eval.}\\ 
\midrule
\textbf{\textit{Document AI Benchmarks}} \\
\quad FUNSD~\cite{jaume2019funsd}        &  \cmark  &  -  &  \xmark  &  \xmark  &  \cmark  &  \xmark   \\
\quad CORD~\cite{park2019cord}           &  \cmark  &  -  &  \xmark  &  \xmark  &  \cmark  &  \xmark   \\
\quad SROIE~\cite{huang2019icdar2019}    &  \cmark  &  -  &  \xmark  &  \xmark  &  \cmark  &  \xmark   \\
\quad VRDU~\cite{wang2023vrdu}           &  \cmark  &  -  &  \xmark  &  \xmark  &  \cmark  &  \xmark   \\
\quad DocVQA~\cite{mathew2021docvqa}     &  \cmark  &  -  &  \xmark  &  \xmark  &  \cmark  &  \xmark   \\
\midrule
\textbf{\textit{Language Agent Benchmarks}} \\
\quad ALFWorld~\cite{shridhar2020alfworld} &  \xmark  &  9  &  \cmark  &  \xmark  &  \cmark  &  \xmark   \\
\quad WebShop~\cite{yao2022webshop} &  \xmark  &  8  &  \cmark  &  \xmark  &  \cmark  &  \xmark   \\
\quad ScienceWorld~\cite{wang2022scienceworld} &  \xmark  &  25  &  \cmark  &  \xmark  &  \cmark  &  \xmark   \\
\quad InterCode~\cite{yang2024intercode} &  \xmark  &  1  &  \cmark  &  \xmark  &  \xmark  &  \xmark   \\
\quad Mind2Web~\cite{deng2024mind2web}   &  \xmark  &  3  &  \cmark  &  \xmark  &  \cmark  &  \xmark   \\
\quad WebArena~\cite{zhou2023webarena}   &  \xmark  &  12 &  \cmark  &  \xmark  &  \cmark  &  \cmark   \\
\quad WebLINX~\cite{lu2024weblinx}       &  \xmark  &  15 &  \cmark  &  \xmark  &  \cmark  &  \xmark   \\
\quad OSWorld~\cite{xie2024osworld}      &  \cmark  &  13 &  \cmark  &  \cmark  &  \xmark  &  \cmark \\ \midrule
\quad \method                            &  \cmark  &  23 &  \cmark  &  \cmark  &  \cmark  &  \cmark    \\
\bottomrule
\end{tabular}
}
}
\vspace{-3mm}
\caption{\textbf{Comparison with recent benchmarks} in document AI and language agent evaluation. It highlights several key strengths of \method. \method excels in cross-application scenarios (\textbf{Cross-App.}), offering a diverse suite of precisely curated customized evaluation functions for each task (\textbf{\# Customized Task Eval.}). Additionally, it supports a larger action space (\textbf{\# Supp. Actions}) and provides more extensible task annotation and environment creation capabilities (\textbf{Extensible}).
}
\label{tab:benchmark-comparisonn}
\end{table*}

%% file: tables/app_action_info_details.tex
\begin{table*}[th]
  \centering
  \small
  \resizebox{\linewidth}{!}{
    \setlength{\tabcolsep}{1mm}{
\begin{tabular}{llll}
\toprule
\textbf{Application} & \textbf{Operations} & \textbf{Arguments} & \textbf{Explanation} \\
\midrule
\multirow{2}[2]{*}{\texttt{System}} & \texttt{switch\_app} & \texttt{target\_app} & Switch to the target application \\
      & \texttt{submit} & \texttt{None} & Finish the operation and submit the results \\
\midrule
\multirow{4}[2]{*}{\texttt{Word}} & \texttt{create\_new\_file} & \texttt{new\_file\_path} & Create a new empty doc file \\
      & \texttt{convert\_to\_pdf} & \texttt{doc\_file\_path, pdf\_file\_path} & Convert a given doc file into a pdf file \\
      & \texttt{read\_doc\_file} & \texttt{file\_path} & Read the contents of a doc file \\
      & \texttt{write\_to\_file} & \texttt{file\_path, contents} & Write text to the doc file \\
\midrule
\multirow{4}[2]{*}{\texttt{Excel}} 
      & \texttt{create\_new\_file} & \texttt{new\_file\_path} & Create a new empty excel file \\
      & \texttt{set\_cell\_content} & \texttt{file\_path, cell\_index, content} & Set a specified cell value in an excel file \\
      & \texttt{delete\_cell\_content} & \texttt{file\_path, cell\_index} & Delete a specified cell in an excel file \\
      & \texttt{read\_excel\_file} & \texttt{file\_path} & Read the contents of an excel file \\
      & \texttt{convert\_to\_pdf} & \texttt{excel\_file\_path, pdf\_file\_path} & Convert a given excel file into a pdf file \\
\midrule
\multirow{4}[2]{*}{\texttt{PDF}} & \texttt{convert\_to\_image} & \texttt{pdf\_file\_path, image\_file\_path} & Convert a given pdf file into an image \\
      & \texttt{convert\_to\_doc} & \texttt{pdf\_file\_path, doc\_file\_path} & Convert a given pdf file into a doc file \\
      & \texttt{read\_pdf\_file} & \texttt{file\_path} & Read the contents of a pdf file with a PDFParser \\
\midrule
\multirow{3}[2]{*}{\texttt{Calendar}} & \texttt{create\_event} & \texttt{username, event\_info} & Create a new calendar event to the specified user \\
      & \texttt{delete\_event} & \texttt{username, event\_id} & Delete a calendar event for the specified user \\
      & \texttt{list\_event} & \texttt{username} & List all the calendar events for the specified user \\
\midrule
\multirow{3}[2]{*}{\texttt{Email}} & \texttt{list\_emails} & \texttt{username} & List all the emails for the specified user \\
      & \texttt{read\_email} & \texttt{username, email\_id} & Read a specified email for the user \\
      & \texttt{send\_email} & \texttt{sender, receiver, email\_contents} & Send an email from one user to another one \\
\midrule
\texttt{OCR} & \texttt{recognize\_text} & \texttt{image\_file\_path} & Use an OCR engine to recognize the text in an image \\
\midrule
\texttt{ChatGPT} & \texttt{query\_chatgpt} & \texttt{query} & Submit a query to ChatGPT \\
\midrule
\texttt{Shell} & \texttt{run\_command} & \texttt{shell\_command} & Run a shell command \\
\bottomrule
\end{tabular}
    }
    }
    \vspace{-3mm}
\caption{\textbf{Applications and their corresponding operations} implemented in \method.}
  \label{tab:app-action-info-detail}%
\end{table*}%

%% file: tables/evaluation_methods.tex

\begin{table*}[htbp]
  \centering
    \resizebox{\linewidth}{!}{
    \setlength{\tabcolsep}{0.8mm}{
    \small
\begin{tabular}{lll}
\toprule
\textbf{Type} & \textbf{Task Examples} & \textbf{Evaluation Functions} \\
\midrule
\multirow{2}[4]{*}{Exact} & Change Carol's midterm1 score to 98 in score excel file & \texttt{excel\_cell\_value}(\texttt{index=}(21,2), \texttt{content=}"98") \\
\cmidrule{2-3}      & Add a paragraph "Approved!" to the end of Application.docx. & \texttt{exact\_match}( \texttt{reference=}"application\_w\_para.docx") \\
\midrule
\multirow{7}[8]{*}{Fuzzy} & Add a meeting to Bob's calendar at 10:30 a.m to 11:00 a.m on & \texttt{contain\_text}(\texttt{text=}["DTSTART:20240517T1030",  \\
& 5/17/2024.     & \quad  "DTEND:20240517T1100", "meeting"]) \\
\cmidrule{2-3}      & Check car trading records and only copy the information about  & \texttt{contain\_text}(\texttt{text=}"Civic") \&\&  \\
      &  my car into car\_records.xlsx, skipping other cars.   & \quad\texttt{not\_contain\_text}(\texttt{text=}"BMW") \\
\cmidrule{2-3}      & Summarize content from the notification image into one notifi-  & 
\multirow{2}[2]{*}{\texttt{file\_exist}(\texttt{file=}"notification.pdf")}
\\      & cation pdf file named notification.pdf. &  \\
\cmidrule{2-3}      & Delete the result files from last month. & \texttt{file\_not\_exist}(\texttt{file=}"April.docx") \\
\midrule
\multirow{4}[4]{*}{Exec.} & \multirow{2}[2]{*}{Find a common time for Bob and Tom for dinner at 5/1/2024.} & \texttt{no\_overlap}("Bob.ics") \&\& \texttt{no\_overlap}("Tom.ics") \&\& \\
      &       &  \quad\texttt{common\_event}("Bob.ics", "Tom.ics", \texttt{event}="dinner") \\

\cmidrule{2-3}      & Randomly assign each student to class 1 to 5 in class member & \texttt{excel\_cell\_comparator}(\texttt{index=}(2,2),  \\
      &  excel file.     & \quad \texttt{comparator=}"lambda x: x in [`1', `2', `3', `4', `5']") \\
\bottomrule
\end{tabular}%
    }
    }
    \vspace{-3mm}
  \caption{\textbf{Evaluation methods and task examples in \method}. We design three types of evaluation methods, Exact Matching, Fuzzy Matching, and Execution-based Evaluation to accurately validate the results of the LLM agents. We skip a few arguments in the evaluation functions due to space limitation.}
  \label{tab:evaluation-method}%
\end{table*}%

%% file: prompts/first_app_switch.tex
\begin{figure*}[ht]
\begin{tcolorbox}[left=1mm,right=1mm,top=0.mm, bottom=0mm,colback=white]
\begin{lstlisting}[style=demo]
========================================================= System ========================================================= 
Today is 2020-05-01 (Friday). The current time is 10:00 AM. You are an AI assistant for user Bob.
You can help solve the task step by step.
You can interact with an operation system and use apps to solve the task.
You must follow the instructions and use the given json format to call APIs.
You can only generate one action at a time.
You can find files for your task in `/testbed/data`. 
You have following apps installed in the system:
 - calendar: an app to manage daily events on calendar.
 - excel: an app to manipulate excel files, including reading, writing, etc.
 - ocr: an app to recognize text from images.
 - pdf: an app to manipulate pdf files, including format conversion and file reading.
 - shell: an app to run shell commands in the system.
 - word: an app to manipulate word files, including reading, writing, converting, etc.
 - email: an app to manage emails, such as sending and reading emails.
 - llm: an app to interact with the large language model to answer questions, generate text, etc.

========================================================= Prompt =========================================================
##Task: Add a meeting to Bob's calendar at 5/17/2024 10:30 a.m to 11:00 a.m
##Available apps: ['calendar', 'excel', 'ocr', 'pdf', 'shell', 'word', 'email', 'llm']
##Instruction:
 - choose an app from the avaiblable apps: {'app': 'system', 'action': 'switch_app', 'target_app': [THE_APP_YOU_CHOOSE]}
##Command:
 
======================================================= Completion =======================================================
```json
{
  "app": "system",
  "action": "switch_app",
  "target_app": "calendar"
}
```
\end{lstlisting}
\end{tcolorbox}
\caption{\textbf{Prompt for application switching} used in \method}
\label{fig:prompt-app-switch}
\end{figure*}

%% file: prompts/mid_prompt.tex
\begin{figure*}[ht]
\begin{tcolorbox}[left=1mm,right=1mm,top=0.mm, bottom=0mm,colback=white]
\begin{lstlisting}[style=demo]
========================================================= System ========================================================= 
Today is 2020-05-01 (Friday). The current time is 10:00 AM. You are an AI assistant for user Bob.
You can help solve the task step by step.
You can interact with an operation system and use apps to solve the task.
You must follow the instructions and use the given json format to call APIs.
You can only generate one action at a time.
You can find files for your task in `/testbed/data`. 
You have following apps installed in the system:
 - calendar: an app to manage daily events on calendar.
 - excel: an app to manipulate excel files, including reading, writing, etc.
 - ocr: an app to recognize text from images.
 - pdf: an app to manipulate pdf files, including format conversion and file reading.
 - shell: an app to run shell commands in the system.
 - word: an app to manipulate word files, including reading, writing, converting, etc.
 - email: an app to manage emails, such as sending and reading emails.
 - llm: an app to interact with the large language model to answer questions, generate text, etc.

========================================================= Prompt =========================================================
##Task: Add a meeting to Bob's calendar at 5/17/2024 10:30 a.m to 11:00 a.m
##History:
 - Step 0: {'app': 'system', 'action': 'switch_app', 'target_app': 'calendar'} -> [Successfully switched to app: calendar]
 - Step 1: {'app': 'calendar', 'action': 'create_event', 'user': 'Bob', 'summary': 'Meeting', 'time_start': '2024-05-17 10:30:00', 'time_end': '2024-05-17 11:00:00'} -> [Successfully create a new event to Bob's calendar.]
##Current apps: calendar
##Instruction: Choose one action from the list as the next step.
 - create a new event to a user's calendar where the time format is '%Y-%m-%d %H:%M:%S':{'app': 'calendar', 'action': 'create_event', 'user': [USER_NAME], 'summary': [EVENT_SUMMARY], 'time_start': [EVENT_START_TIME], 'time_end': [EVENT_END_TIME]}
 - delete an event from a user's calendar given the event summary:{'app': 'calendar', 'action': 'create_event', 'user': [USER_NAME], 'summary': [EVENT_SUMMARY]}
 - list all events from a user's calendar: {'app': 'calendar', 'action': 'list_events', 'username': [USER_NAME]}
 - switch to another app among ['excel', 'ocr', 'pdf', 'shell', 'word', 'email', 'llm']: {'app': 'system', 'action': 'switch_app', 'target_app': [THE_APP_YOU_CHOOSE]}
 - finish the task with your answer as None if the task is not a question: {'app': 'system', 'action': 'finish_task', 'answer': 'None'}
 - finish the task with your answer if the task is a question: {'app': 'system', 'action': 'finish_task', 'answer': [ANSWER]}
##Command:
 
======================================================= Completion =======================================================
{'app': 'system', 'action': 'finish_task', 'answer': 'None'}
\end{lstlisting}
\end{tcolorbox}
\caption{\textbf{Prompt for planning next operation based on the trajectory} used in \method}
\label{fig:prompt-plan}
\end{figure*}

%% file: prompts/ablation_prompt.tex
\begin{figure*}[ht]
\begin{tcolorbox}[left=1mm,right=1mm,top=0.mm, bottom=0mm,colback=white]
\begin{lstlisting}[style=demo]
========================================================= System ========================================================= 
Today is 2020-05-01 (Friday). The current time is 10:00 AM. You are an AI assistant for user Bob.
You can help solve the task step by step.
You can interact with an operation system and use apps to solve the task.
You must follow the instructions and use the given json format to call APIs.
You can only generate one action at a time.
You can find files for your task in `/testbed/data`. 
You have following apps installed in the system:
 - calendar: an app to manage daily events on calendar.
 - excel: an app to manipulate excel files, including reading, writing, etc.
 - ocr: an app to recognize text from images.
 - pdf: an app to manipulate pdf files, including format conversion and file reading.
 - shell: an app to run shell commands in the system.
 - word: an app to manipulate word files, including reading, writing, converting, etc.
 - email: an app to manage emails, such as sending and reading emails.
 - llm: an app to interact with the large language model to answer questions, generate text, etc.

========================================================= Prompt =========================================================
##Task: Add a meeting to Bob's calendar at 5/17/2024 10:30 a.m to 11:00 a.m
##History:
 - Step 0: {'app': 'calendar', 'action': 'create_event', 'user': 'Bob', 'summary': 'Meeting', 'time_start': '2024-05-17 10:30:00', 'time_end': '2024-05-17 11:00:00'} -> [Successfully create a new event to Bob's calendar.]
##Instruction: Choose one action from the list as the next step.
 - create a new event to a user's calendar where the time format is '%Y-%m-%d %H:%M:%S':{'app': 'calendar', 'action': 'create_event', 'user': [USER_NAME], 'summary': [EVENT_SUMMARY], 'time_start': [EVENT_START_TIME], 'time_end': [EVENT_END_TIME]}
 - delete an event from a user's calendar given the event summary:{'app': 'calendar', 'action': 'delete_event', 'user': [USER_NAME], 'summary': [EVENT_SUMMARY]}
 - list all events from a user's calendar: {'app': 'calendar', 'action': 'list_events', 'username': [USER_NAME]}
 - read the excel file to see the existing contents: {'app': 'excel', 'action': 'read_file', 'file_path': [THE_PATH_TO_THE_EXCEL_FILE]
 - write text to a cell in the excel file (index starts from 1): {'app': 'excel', 'action': 'set_cell', 'file_path': [THE_PATH_TO_THE_EXCEL_FILE], 'row_idx': [THE_ROW_INDEX], 'column_idx': [THE_COLUMN_INDEX], 'text': [THE_TEXT_TO_WRITE]}
 - delete text in a cell of the excel file (index starts from 1, delete means set empty): {'app': 'excel', 'action': 'delete_cell', 'file_path': [THE_PATH_TO_THE_EXCEL_FILE], 'row_idx': [THE_ROW_INDEX], 'column_idx': [THE_COLUMN_INDEX]
 - create a new excel file: {'app': 'excel', 'action': 'create_new_file', 'file_path': [THE_PATH_TO_THE_NEW_EXCEL_FILE]}
 - convert an excel document to a pdf: {'app': 'excel', 'action': 'convert_to_pdf', 'excel_file_path': [THE_PATH_TO_THE_EXCEL_FILE], 'pdf_file_path': [THE_PATH_TO_THE_PDF_FILE]}
 - recognize the text from an image file: {'app': 'ocr', 'action': 'recognize_file', 'file_path': [THE_PATH_TO_THE_IMAGE_FILE]}
 - convert a pdf file to an image file: {'app': 'pdf', 'action': 'convert_to_image', 'pdf_file_path': [THE_PATH_TO_THE_PDF_FILE], 'image_file_path': [THE_PATH_TO_THE_IMAGE_FILE]}
 - convert a pdf file to a word file: {'app': 'pdf', 'action': 'convert_to_word', 'pdf_file_path': [THE_PATH_TO_THE_PDF_FILE], 'word_file_path': [THE_PATH_TO_THE_WORD_FILE]}
 - read a pdf file: {'app': 'pdf', 'action': 'read_file', 'pdf_file_path': [THE_PATH_TO_THE_PDF_FILE]}
 - Send an email to a recipient: {'app': 'email', 'action': 'send_email', 'sender': [SENDER], 'recipient': [RECIPIENT], 'subject': [SUBJECT], 'content': [CONTENT]}
 - List emails for a given username: {'app': 'email', 'action': 'list_emails', 'username': [USER_NAME]}
 - Read a user's email by the given Email ID: {'app': 'email', 'action': 'read_email', 'username': [USERNAME], 'email_id': [EMAIL_ID]}
 - run a shell command: {'app': 'shell', 'action': 'command', 'command': [THE_COMMAND_YOU_WISH_TO_RUN]}
 - convert a word document to a pdf: {'app': 'word', 'action': 'convert_to_pdf', 'word_file_path': [THE_PATH_TO_THE_WORD_FILE], 'pdf_file_path': [THE_PATH_TO_THE_PDF_FILE]}
 - create a new word file: {'app': 'word', 'action': 'create_new_file', 'file_path': [THE_PATH_TO_THE_NEW_WORD_FILE]}
 - read the content of the word file: {'app': 'word', 'action': 'read_file', 'file_path': [THE_PATH_TO_THE_WORD_FILE]
 - write text to a word file: {'app': 'word', 'action': 'write_to_file', 'file_path': [THE_PATH_TO_THE_WORD_FILE], 'contents': [THE_CONTENTS_YOU_WISH_TO_WRITE]}
 - Query an LLM model for an answer to a given prompt: {'app': 'llm', 'action': 'complete_text', 'prompt': [PROMPT]}
 - finish the task with your answer as None if the task is not a question: {'app': 'system', 'action': 'finish_task', 'answer': 'None'}
 - finish the task with your answer if the task is a question: {'app': 'system', 'action': 'finish_task', 'answer': [ANSWER]}
##Command:
 
======================================================= Completion =======================================================
```json
{
  "app": "system",
  "action": "finish_task",
  "answer": "None"
}
```
\end{lstlisting}
\end{tcolorbox}
\caption{\textbf{Prompt of \textit{List All Operations} used in the ablation study}}
\label{fig:prompt-ablation}
\end{figure*}